\documentclass{article}

\usepackage{makecell}
\usepackage{mdframed}
\usepackage{microtype}
\usepackage{graphicx}
\usepackage{subfigure}
\usepackage{booktabs} 

\usepackage{hyperref}

\usepackage[accepted]{icml2025} 

\usepackage{amsmath}
\usepackage{amssymb}
\usepackage{mathtools}
\usepackage{amsthm}

\usepackage[capitalize,noabbrev]{cleveref}

\theoremstyle{plain}

\theoremstyle{definition}

\theoremstyle{remark}

\usepackage[textsize=tiny]{todonotes}

\icmltitlerunning{Automatically Interpreting Millions of Features in Large Language Models}

\begin{document}

\twocolumn[
\icmltitle{Automatically Interpreting Millions of Features in Large Language Models}



\icmlsetsymbol{equal}{*}


\begin{icmlauthorlist}
\icmlauthor{Gonçalo Paulo}{eai}
\icmlauthor{Alex Mallen}{eai}
\icmlauthor{Caden Juang}{nw}
\icmlauthor{Nora Belrose}{eai}
\end{icmlauthorlist}

\icmlaffiliation{eai}{EleutherAI}
\icmlaffiliation{nw}{Northwestern University, Evanston, Illinois, USA}

\icmlcorrespondingauthor{Gonçalo Paulo}{goncalo@eleuther.ai}

\icmlkeywords{Machine Learning, ICML}

\vskip 0.3in
]



\printAffiliationsAndNotice{}  

\begin{abstract}
While the activations of neurons in deep neural networks usually do not have a simple human-understandable interpretation, sparse autoencoders (SAEs) can be used to transform these activations into a higher-dimensional latent space which can be more easily interpretable. However, SAEs can have millions of distinct features, making it infeasible for humans to manually interpret each one. In this work, we build an open-source automated pipeline to generate and evaluate natural language interpretations for SAE features using LLMs. We test our framework on SAEs of varying sizes, activation functions, and losses, trained on two different open-weight LLMs. We introduce five new techniques to score the quality of interpretations that are cheaper to run than the previous state of the art. One of these techniques, intervention scoring, evaluates the interpretability of the effects of intervening on a feature, which we find explains features that are not recalled by existing methods. We propose guidelines for generating better interpretations that remain valid for a broader set of activating contexts, and discuss pitfalls with existing scoring techniques.
Our code is available \href{https://github.com/EleutherAI/delphi}{online}.
\end{abstract}

\section{Introduction}

Large language models (LLMs) have reached human level performance in a broad range of domains \citep{gpt4report}, and can even be leveraged to develop agents \citep{voyager_agent} that can strategize \citep{Bakhtin2022HumanlevelPI}, cooperate and develop new ideas \citep{ai_scientist,maia}. At the same time, we understand little about the internal representations driving their behavior. Early mechanistic interpretability research focused on analyzing the activation patterns of individual neurons \citep{Olah2020,gurnee2023finding,gurnee2024universal}. Due to the large number of neurons to be interpreted, automated approaches were proposed \citep{bills2023language}, where a second LLM is used to propose an interpretation given a set of neuron activations and the text snippets it activates on, in a process similar to that of generating a human label. But research has found that most neurons are ``polysemantic'', activating in contexts that can be significantly different \citep{arora2018linear,elhage2022toy}. The Linear Representation Hypothesis \citep{Park2023TheLR} posits that human-interpretable concepts are encoded in \emph{linear combinations} of neurons. A significant branch of current interpretability work focuses on extracting these features and disentangling them \citep{bereska2024mechanistic}.

Sparse autoencoders (SAEs) were proposed as a way to address polysemanticity \citep{cunningham2023sparse}. SAEs consist of two parts: an encoder that transforms activation vectors into a sparse, higher-dimensional latent space, and a decoder that projects the features back into the original space. Both parts are trained jointly to minimize reconstruction error.  SAE features\footnote{We use feature to mean the intermediate dimension of the SAE like in \citep{bricken2023monosemanticity}.} were found to be interpretable and potentially more monosemantic than neurons \citep{bricken2023monosemanticity, cunningham2023sparse}. Recently, a significant effort was made to scale SAE training to larger models, like GPT-4 \citep{gao2024scaling} and Claude 3 Sonnet \citep{templeton2024scaling}, and they have become an important interpretability tool for LLMs. 

Training an SAE yields numerous sparse features, each of which needs a natural language interpretation. In this work, we introduce an automated framework that uses LLMs to generate an interpretation for each feature in an SAE. We use this framework to explain millions of features across multiple models, layers, and SAE architectures. We also propose new ways to evaluate the quality of interpretations, and discuss the problems with existing approaches. We hope that these result can inform other repositories of interpretations for SAE features that already exist \citep{neuronpedia}, as well as improve comparisons between different SAE architectures.

\section{Related Work}
One of the most successful approaches to automated interpretability focused on explaining neurons of GPT-2 using GPT-4 \citep{bills2023language}. GPT-4 was shown examples of contexts where a given neuron was active and was tasked to provide a short interpretation that could capture the activation patterns. To evaluate if a given interpretation captured the behavior of the neuron, GPT-4 was tasked to predict the activations of the neuron in a given context having access to that interpretation. The interpretation is then scored by how much the simulated activations correlate with the true activations. A similar approach was used in \citet{templeton2024scaling} to explain SAE features of Claude 3 Sonnet. In general, current approaches focus on collecting contexts together with feature activations from the model to be explained, and use a larger model to find patterns in activating contexts.

Following those works, other methods of evaluating interpretations have been proposed, including asking a model to generate activating examples and measuring how much a "module" activates \citep{singh2023explaining,cosy}. More recently, ``interpretability agents" have been built, managing to do iterative experiment to find the best interpretations of vision neurons \citep{maia}. Potentially cheaper versions of automated interpretability have been suggested, where the model that is being explained doubles as an interpretation generation model \citep{kharlapenko-2024}. A prompt querying the meaning of a single placeholder token is passed to the model and feature activations are patched into its residual stream at the position of the placeholder token during execution, generating continuations related to the feature. This technique is inspired by earlier work on Patchscopes \citep{patchscopes} and SelfIE \citep{selfie}.

\section{Automated interpretability pipeline}

\begin{figure*}[h]
    \centering
    \includegraphics[width=1\linewidth]{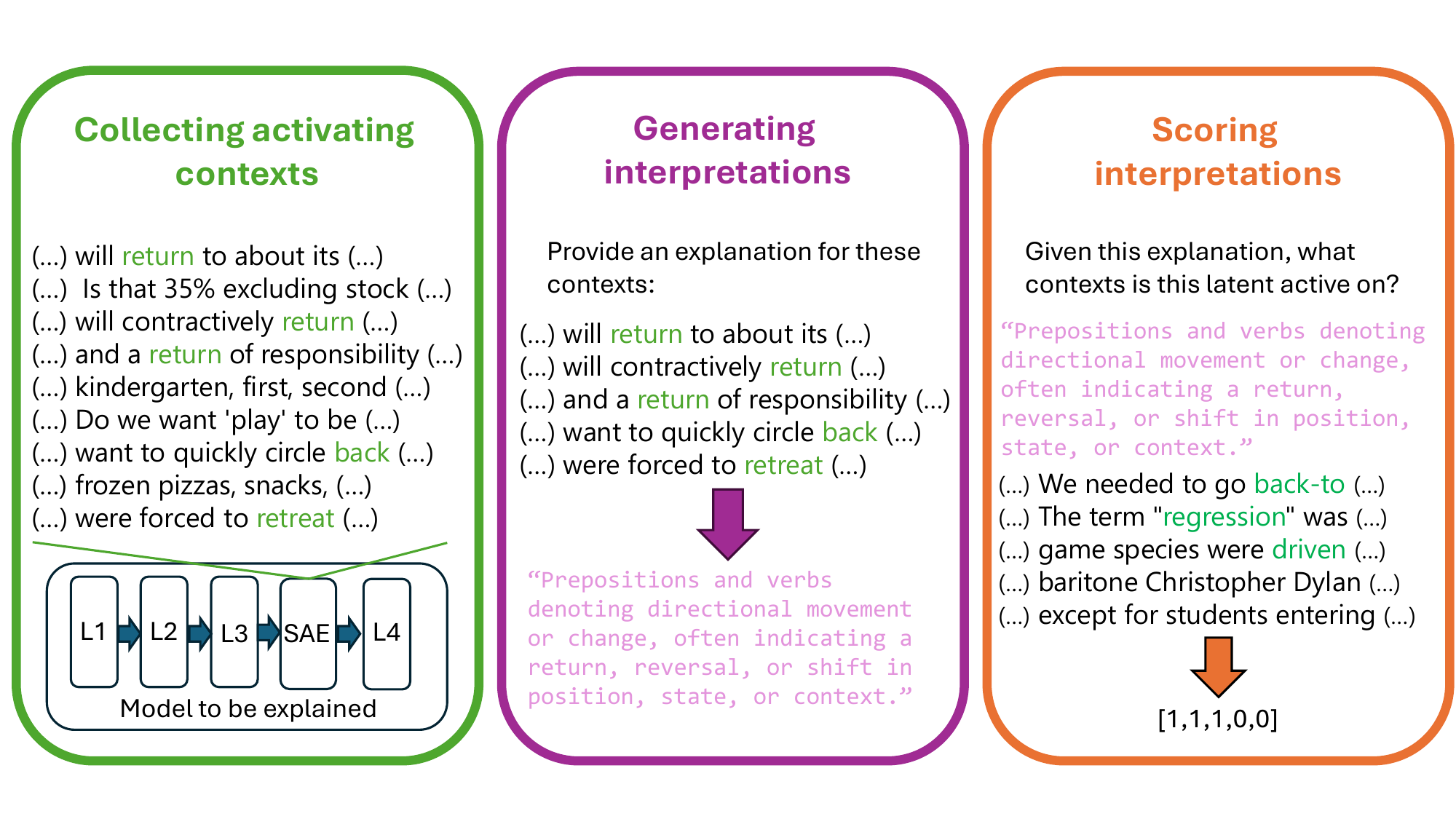}
    \vskip -0.7cm
    \caption{\textbf{Auto-interpretability pipeline.} The first step of the interpretability pipeline is collecting the activations of the target SAE over a broad range of text. In this figure we show short contexts for legibility, and over represent the active feature, which in reality would be active in a very small portion of the text. The activating contexts are selected and shown to an explainer LLM, prompt in Appendix \ref{app:explainer-prompt}, which provides a short interpretation. This interpretation is then given to a scorer LLM that is tasked to use this interpretation to distinguish activating from non activating contexts, see Section \ref{sec:scoring_methods} }
    \label{fig:pipeline}
\end{figure*}

In this section we explain, step-by-step, the main pipeline used to generate interpretations and score them. The pipeline can be broadly divided into 3 sequential steps. First, the activations of the SAEs to be interpreted are collected over a broad dataset. Then, for each feature, relevant contexts are selected and shown to an LLM which generates an interpretation for the activation pattern observed. Finally, these interpretations are matched with different contexts and used by an LLM in different tasks that evaluate how good the interpretations are in predicting activating and non-activating contexts, details explained below. As an illustration, we represent how that pipeline might look like for a real feature, see Figure \ref{fig:pipeline}. 

\subsection{Collecting activations}

Sparse autoencoders (SAEs) are feedforward networks with one hidden layer trained to reproduce their input using a sparse number of neurons. This means that while there are several times more neurons than there are dimensions of the input, only a small fraction of these neurons-- less than 60 in this work-- are non-zero and contribute to the output. In our pipeline, we seek to explain the non-zero activations of the SAE hidden layer neurons, which we call ``features.''

We collected feature activations from the SAEs over a 10M token sample of RedPajama-v2 (RPJv2; ~\citealt{together2023redpajama}). This is the same dataset that we used to train the Llama 3.1 8b SAEs, and consists of a data mix similar to the Llama 1 pretraining corpus. 

We collected batches of 256 tokens starting with the beginning of sentence token (BOS).\footnote{We throw out the activations on the BOS token when generating interpretations for Gemma, as we were told in personal communication these SAEs were not trained on BOS activations.} The contexts used for activation collection are smaller than the contexts used to train the SAEs, and we find that, on average, 30\% of the features, per layer, of the 131k feature Gemma 2 9b SAE don't activate more than 200 times over these 10M tokens and 15\% don't activate at all. When we consider the training context length of 1024, only 5\% of features don't fire. When using a closer proxy of the training data, the \href{https://huggingface.co/datasets/monology/pile-uncopyrighted}{``un-copyrighted'' Pile}, we find that the number of features that activate fewer than 200 times decreases to 15\%, and that only 1\% don't activate at all, even when considering contexts of size 256. Interestingly, in the case of the 16k feature Gemma 2 9b SAE only around 10\% of features activate fewer than 200 times on the RPJv2 sample, suggesting that the larger 131k feature SAE learned more dataset specific features than its smaller cousin.


\subsection{Generating interpretations}

Our approach to generating interpretations follows \citet{bricken2023monosemanticity} in showing the explainer model examples sampled from different quantiles, but uses a more ``natural'' prompt where the activating example is shown whole, with the activating tokens emphasized and their activation strength shown after the example, see Appendix~\ref{app:explainer-prompt} for the full prompt, similar to what is done in \citep{prompt_neurons}. For examples of some interpretations and their scores, see Figure \ref{fig:5Mfeatures}.  

We show the explainer model, Llama 3.1 70b Instruct, 40 different activating examples, each activating example consisting of 32 tokens, where the activating tokens can be on any position. While most features are well-explained by contexts of length 32, we expect features active over longer contexts to not be well captured by our pipeline. Using longer contexts to generate explanations would require either showing the model less examples, or using a much more powerful (and expensive) explainer model. Because we are able to evaluate interpretations significantly cheaper, bad interpretation can easily be filtered out, and we leave the explanation of ``long-range'' features for future work. Using this interpretation setup would cost  \$200 to interpret 1 million features.

\begin{figure*}[h!]
    \centering
    \includegraphics[width=1\linewidth]{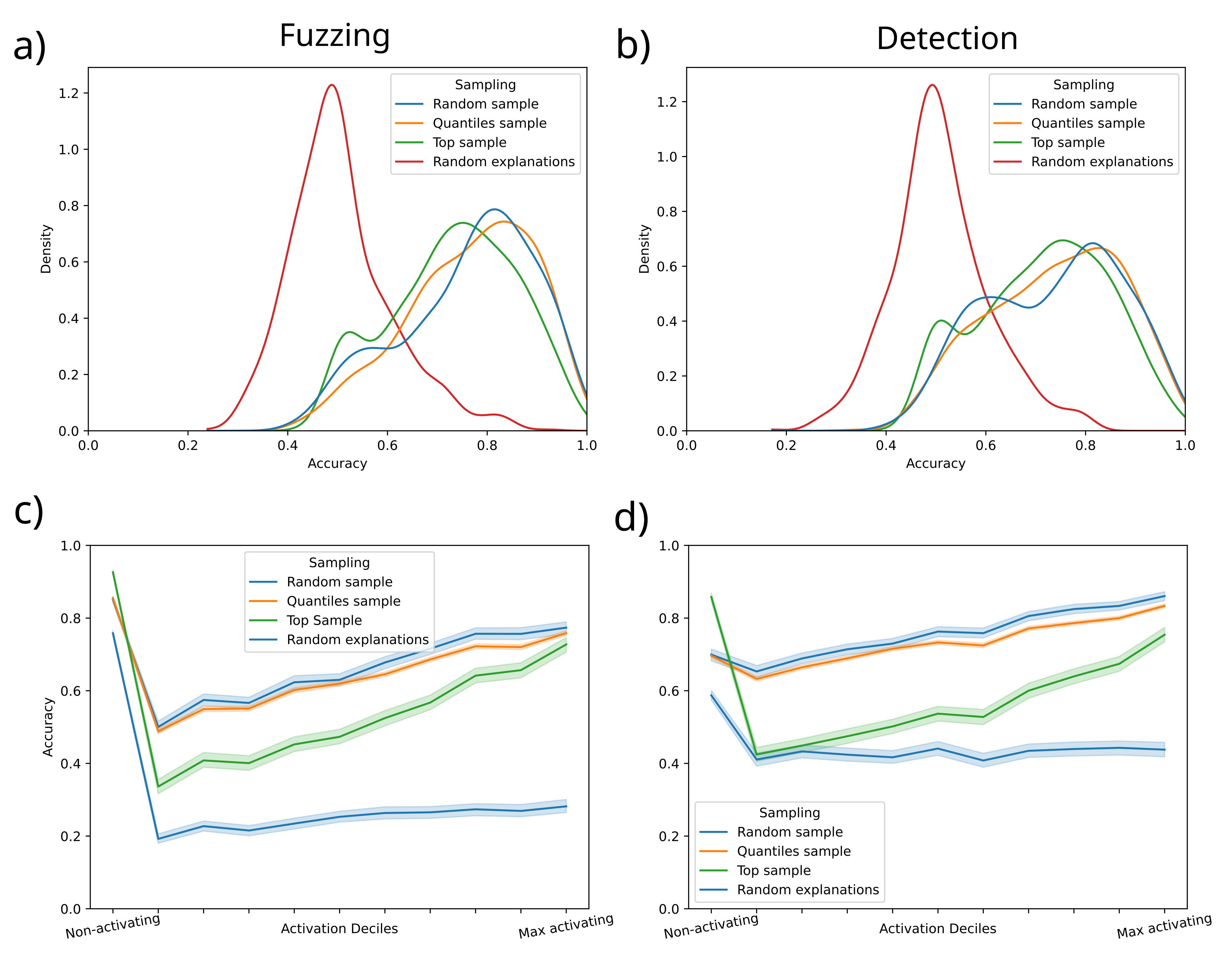}
    \vskip -0.2cm
    \caption{\textbf{Fuzzing and detection scores for different sampling techniques.} Panels \textbf{a)} and \textbf{b)} show the distributions of fuzzing and detection scores, respectively, as a function of different example sampling methods for interpretation generation. Sampling only from the top activation gets on average lower accuracy in fuzzing and on detection when compared with uniform sampling and stratified sampling. The distributions from random sampling and sampling from quantiles are very similar. Panels \textbf{c)} and \textbf{d)} measure how the interpretations generalize across activation quantiles, showing that interpretations generated from the top quantiles are better at distinguishing non-activating examples, but have lower accuracy on other quantiles, especially on the lower activating deciles. This also happens for the other interpretations, but the accuracy does not drop as much in lower activating deciles. We also show the scores of random explanations.}
    \label{fig:accuracy}
\end{figure*}

We find that randomly sampling from a broader set of examples leads to interpretations that cover a more diverse set of activating examples, sometimes to the detriment of the top activating examples, see Figure \ref{fig:accuracy}. Using only top activating examples often yields more concise and specific interpretations that accurately describe these examples, but which fail to capture the whole distribution. On the other hand, stratified sampling from the deciles of the activation distribution can lead to interpretations that are too broad and that fail to capture a meaningful interpretation. For examples of these types of failure modes, see Figure \ref{fig:top_vs_random} and the discussion in the Appendix \ref{app:examples_contexts}. 

\subsubsection{Automatically interpreting interventions}
\label{sec:output}

Automatic interpretability methods, including most of those explored in this paper, typically look for correlations between activation of a feature and some natural-language property of the input. However, some features are more closely related to what the model will output. For example, we find a feature\footnote{feature 157 of Gemma 2 9b's layer 32 autoencoder with 131k features and an average L0 norm of 51. Its detection score is 0.6.} whose activation causes the model to output words associated with reputation but does not have a simple interpretation in terms of inputs.

We define an \emph{output feature} as a feature that causes a property of the model's output that can be easily explained in natural language. See Section~\ref{sec:output-scoring} for the definition we use in scoring.

Output features can also be described in terms of their correlation with inputs. For example, the ``reputation'' feature activates in contexts where likely next tokens relate to reputation. However, explaining output features in terms of causal influence on output has two advantages.

\begin{enumerate}
    \item \textbf{Scalability}. Output features are easier to describe in terms of effect on output because the explainer only needs to notice a simple pattern of output. The pattern of inputs this feature correlates with is more complex. Explaining a feature by correlating it with inputs requires approximating the computation of the subject model leading up to that feature, which may be challenging when subject models are highly capable and performing difficult tasks. Some features' influence on output, however, might remain easily explainable.
    \item \textbf{Causal evidence}. We might like to know that a feature causes a property of the model's output so that we can steer the model. Previous work has shown that existing auto-interpretability scores fail to accurately capture how well a given interpretation can predict the effect of intervening on a given neuron \citep{huang2023rigorously}. Further, prior work has argued that causal evidence is more robust to distribution shifts~\citep{buhlmann2018invariance, scholkopf2012causal}.
\end{enumerate}

\subsection{Scoring interpretations}
\label{sec:scoring_methods}

Not only should the generation of interpretations efficient, but so should the scoring of such interpretations. Practitioners would like to know how faithful each interpretation is to the actual behavior of the network. Some features may not be amenable to a simple interpretation, and in these cases we expect the auto-interpretability pipeline to output an interpretation with poor evaluation metrics. Secondly, we can use evaluations as a feedback signal for the training of explainer models and tuning hyperparameters in the pipeline. Finally, some SAEs may simply be poor quality overall, and the aggregate evaluation metrics of the SAE's interpretations can be used to detect this.

The quality of SAE interpretations has so far been measured via simulation scoring, a technique introduced by \citet{bills2023language} for evaluating interpretations of neurons. It involves asking an explainer model to ``simulate'' a feature in a set of contexts, which means predicting how strongly the feature should activate at each token in a context given an interpretation. The Pearson correlation between the simulated and true activations is the simulation score. The standard in the literature is to sample contexts from a ``top-and-random'' distribution that mixes maximally activating contexts and contexts sampled uniformly at random from the activating corpus. While oversampling the top activating contexts introduces bias, it is used as a cheap variance reduction technique, given that simulation scoring over hundreds of examples per feature would be expensive \citep{cunningham2023sparse}. 

In this work, we take a slightly different view of what makes for a good interpretation: an interpretation should serve as a \emph{binary classifier} distinguishing activating from non-activating contexts. The reasoning behind this is simple. Given the highly sparse nature of SAE features, most of their variance could be captured by a binary predictor that predicts the mean nonzero activation when the feature is expected to be active, and zero otherwise. In statistics, \href{https://en.wikipedia.org/wiki/Zero-inflated_model}{zero-inflated} phenomena are often modeled as a mixture distribution with two components: a Dirac delta on zero, and another distribution (e.g. Poisson) for nonzero values. This goes against the general wisdom of simulation scoring, which focuses only on activating examples, even though they are less than 0.01\% of the relevant contexts for each feature of an SAE and that being able to distinguish non-activating contexts from activating contexts seems to be relatively more important.


As an alternative to simulation scoring, we introduce four new evaluation methods that focus on how well an interpretation enables a scorer to discriminate between activating and non-activating contexts. As an added benefit, all of these methods are more compute-efficient than simulation, see Table \ref{tab:cost}. We also propose a scoring method that evaluates the interpretability of interventions using a specific feature. 
 
\subsubsection{Detection}

One simple approach for scoring interpretations is to ask a language model to identify whether a whole sequence activates a SAE feature given an interpretation. Detection requires few output tokens for each example used in scoring, meaning examples from a wider distribution of feature activation strengths can be used at approximately the same expense. By including non-activating contexts, this method measures both the precision and recall of the interpretation. 

Detection is more ``forgiving'' than simulation insofar as the scorer does not need to localize the feature to a particular token. This means that detection evaluates whether an interpretation correctly identifies the types of contexts that a feature is active on, even if the interpretation does not correctly identify the actual token. 

Both this method and fuzzing (described next) can leverage token probabilities to estimate how certain the scorer model is of their classification, and this is an effect that we believe can be to improve the scoring methods. Details on the prompt in Appendix \ref{app:detection_details}. 

\subsubsection{Fuzzing} 

Fuzzing is similar to detection, but at the level of individual tokens. Here, potentially activating tokens are delimited in each example and the language model is prompted to identify which of the sentences are correctly marked. Fuzzing is the most similar with simulation. Because SAE activations are very sparse, simulation scoring mostly boils down to correctly identifying which of the tokens have non-zero activations. For this reason, fuzzing is the score that most correlates with simulation.

When interpretations focus on which tokens the feature activates, but not on which contexts, they can still score high on fuzzing, but comparatively lower on detection, specially if they are common tokens. Thus, evaluating an interpretation on both detection and fuzzing can identify whether a model is classifying examples for the correct reason. Details on the prompt on Appendix \ref{app:fuzzing_details}.

\subsubsection{Surprisal}

Surprisal scoring is based on the idea that a good interpretation should help a base language model $\mathcal M$ (the ``scorer'') achieve lower cross-entropy loss on activating contexts than it would without a relevant interpretation. Specifically, for each context $\mathbf{x}$, activating or non-activating, we measure the \emph{information value} of an interpretation $\mathbf{z}$ as $\log p_{\mathcal M}(\mathbf{x}|\mathbf{z}) - \log p_{\mathcal M}(\mathbf{x}|\Tilde{\mathbf{z}})$, where $\Tilde{\mathbf{z}}$ is a fixed pseudo-interpretation. A good interpretation should have higher information value on activating examples than on non-activating ones. 

The overall surprisal score of the interpretation is given by the AUROC of its information value when viewed as a classifier distinguishing activating from non-activating contexts. Details on the prompt and how to compute the score in Appendix \ref{app:surprisal_details}. 

Surprisal relies on the ability of the scorer model to tune its predictions based on the explanation, measuring both the relevance of the context given in the interpretation and the token specificity, but we believe that the current setup could be improved, as this is the score with the least correlation with the others.

\subsubsection{Embedding}

Classifying between active and non-active contexts given a certain interpretation can also be seen as using interpretations of features as ``queries'' that should be able to retrieve relevant ``documents'', contexts where the feature is active, between non-relevant ``documents'', non-activating contexts. This way, we take a selection of activating and non-activating contexts that embedded by an encoding transformer, and the similarity between the query and the documents is used as a classifier to distinguish between activating and non-activating contexts, and the score is the AUROC.

If the encoding model is small enough - we used a 400M parameter model - this technique is the fastest and opens up the possibility to evaluate a larger fraction of the activation distribution. We have seen that using a larger embedding model - 7B parameter model - didn't significantly improve the scores, see fig \ref{fig:embedding}, although we believe that this approach was under-investigated. Details on the prompt, on the embedding model and on the way to compute the score in Appendix \ref{app:embedding_details}.

\subsubsection{Intervention scoring}
\label{sec:output-scoring}
Unlike the above four context-based scores, intervention scoring interprets a feature's counterfactual impact on model output. We quantify the interpretability of an intervention $I$ with interpretation $\mathbf  z$ on a distribution of prompts $\pi$ as the average decrease in the scorer's surprisal about the interpretation when conditioned on text generated with the intervention.
\begin{align}
    S = \mathbb E_{\mathbf x}  \Big [ \mathbb E_{\mathbf i \sim \mathcal G_I(\mathbf x)}[\log p_{\mathcal M}(\mathbf z | \mathbf i)] - \mathbb E_{\mathbf g \sim \mathcal G(\mathbf x)}[\log p_{\mathcal M}(\mathbf z | \mathbf g)] \Big] \nonumber
\end{align}
$\mathcal G(\mathbf x)$ is the distribution over subject model generations given prompt $\mathbf x$ at temperature 1. In $\mathcal G_I(\mathbf x)$, the intervention is applied to the subject model as it generates. In practice we estimate the quantity $S$ by sampling one clean and one intervened generation for each sampled of prompt.

Sufficiently strong interventions can be trivially interpretable by causing the model to deterministically output some logit distribution. Therefore, \textbf{interpretability scores of interventions should be compared for interventions of a fixed strength}. We define the strength $\sigma$ of an arbitrary intervention $I$ as the average KL-divergence of the model's intervened logit distribution with reference to the model's clean output.
\begin{equation}
    \sigma(I; \pi) = \mathbb E_{\mathbf x\sim \pi} \big[D_{KL}(p_{\text{subject}}(\cdot|\mathbf x)\ ||\ p_{\text{subject}, I}(\cdot|\mathbf x))\big] \nonumber
\end{equation}
See Appendix~\ref{app:output} for details on the interpretation and scoring pipeline we use in our intervention experiments.

\section{Results}
\subsection{Comparing scoring methods}
\begin{table*}[h]
\caption{Estimated cost of the different scoring methods. Brackets represent tokens that can be cached, potentially saving costs, although we do not do these calculations here. The number of input and output tokens is computed considering 100 examples show, and using fewer examples would cost proportionally less.}
\label{tab:cost}
\centering
\begin{tabular}{lccccc}
\toprule
           & Input tokens & Output tokens & \makecell{Cost (\$/100k features) \\ Llama 70b} & \makecell{Cost (\$/100k features) \\ Claude Sonnet 3.5 }  \\
\midrule
Fuzzing    & 19.6k  (14.2k)      & 249      & 676          &  6.2k \\
Detection  & 17.0k  (11.9k)     & 240         & 588       & 5.5k \\
Simulation (AAO) & 104.9k  (87.5k)  & 5    &  3.6k         & 31.5k              \\
Simulation (TBT) & 496.9k (451k)        &  46.7k        &   18.7k & 219.1k               \\
\bottomrule
\end{tabular}
\end{table*}
When wanting to evaluate the explanation over a larger number of examples, simpler methods than simulation scoring have to be used \citep{templeton2024scaling}. Simulation scoring is traditionally done only in activating examples, ignoring whether the proposed explanation correctly handles non-activating examples. Simulation scoring is normally done in a small amount of examples per feature due to its high cost, and this is a significant disadvantage that prompted us to investigate more efficient scoring methods.

In Table \ref{tab:cost} we compare the amount of tokens used to score a single feature using 100 different contexts, for both fuzzing, detection and two different simulation methods. We can see that fuzzing and detection are at least 5x cheaper than simulation when using the all-at-once (AAO) method described in \cite{bills2023language}, although this method requires access to the log-probabilites of the prompt tokens, which are not accessible from providers of state-of-the-art closed-source models. A researcher wanting to generate simulation scores on the fly, without access to a local model, would have to use the token-by-token (TBT) method, which is 30x more than expensive than both fuzzing and detection. For the same set of intepretations, fuzzing and detection scores given by Llama 70b and Claude Sonnet 3.5 have similar distributions. On the other hand, Claude Sonnet 3.5 produces to higher simulation scores on average than Llama 70b, see Table \ref{tab:model_size} meaning that we should probably compare the price of fuzzing and detection on Llama 70b to the price of simulation using Claude Sonnet 3.5.

Embedding scoring is even cheaper, requiring 4000 input tokens for 100 different contexts, which at an average of \$0.13 per million tokens would only cost \$50/100k features. 

We propose that feature interpretations should be evaluated using more than a single technique when possible, because as discussed in Section \ref{sec:scoring_methods}, each of the proposed methods have different failure modes, and a more rich scoring framework can address some of these flaws. Specifically, we find that when computing the correlation of scores computed on the same 800 explanations, we find that while there is a clear positive correlation, some scores disagree -- see Appendix \ref{app:score_correlation} and Tables \ref{tab:spearman} and \ref{tab:pearson}.

We find fuzzing to be the most correlated with simulation, as fuzzing is mostly measuring how well the explanation can help the model predict which tokens have non-zero activation. Due to the sparsity of SAE activations, simulation scoring is doing a very similar test. On the contrary, we find that simulation has a significantly lower correlation with detection, embedding and surprisal scoring, as they more accurately measure whether a context activates a given feature, instead of which token is active in a given context. When comparing the simulation scores provided by Claude 3.5 Sonnet, all correlations have a slight increase, see Appendix \ref{app:score_correlation}.

We also measure the correlation between our scoring methods and scores given by humans. Humans were tasked to score how an explanation related to a given context, similar to that done in \citep{templeton2024scaling}. The human evaluators could choose between 4 options, ``1 - Explanation not related to the text'', ``2 - Explanation vaguely related to the text'', ``3 - Incomplete explanation but present in the text'', ``4 - Explanation describes the activation of the tokens''. We measure the average score of a feature in at least 5 contexts. In total 700 contexts and 81 feature interpretation. We find that fuzzing as the best Spearman correlation with human scores (0.69), followed by simulation (0.60), and detection (0.59). Surprisal and embedding have 0.34 and 0.32 Spearman correlation respectively. 

Due to their relative cheapness compared with simulation scoring, we propose that fuzzing and detection scoring can be used to estimate whether explanations correctly identify on which tokens a feature is most likely to be active, and in which types of contexts this happens. Cheaper methods like embedding can be used to quickly iterate on explanations or to broadly separate features that have bad explanations from those with good explanations, so that they can be refined. 

\subsubsection{Intervention scoring}

The previously discussed scoring methods are \emph{correlational} in the sense that they measure whether an interpretation can be used to predict the activation values of a given feature, or distinguish between activating and non-activating examples. Intervention scoring proposes to measure how well a given interpretation can predict the effect of \emph{interventions} on the corresponding feature. Here we compare correlational interpretations generated with our pipeline and scored with fuzzing, to a set of interventional interpretations scored with intervention scoring. Our hypothesis is that some features will have low correlational scores because their behavior is better explained by their downstream effects than by the contexts where they are active.

\begin{figure*}
    \centering
    \includegraphics[width=\linewidth]{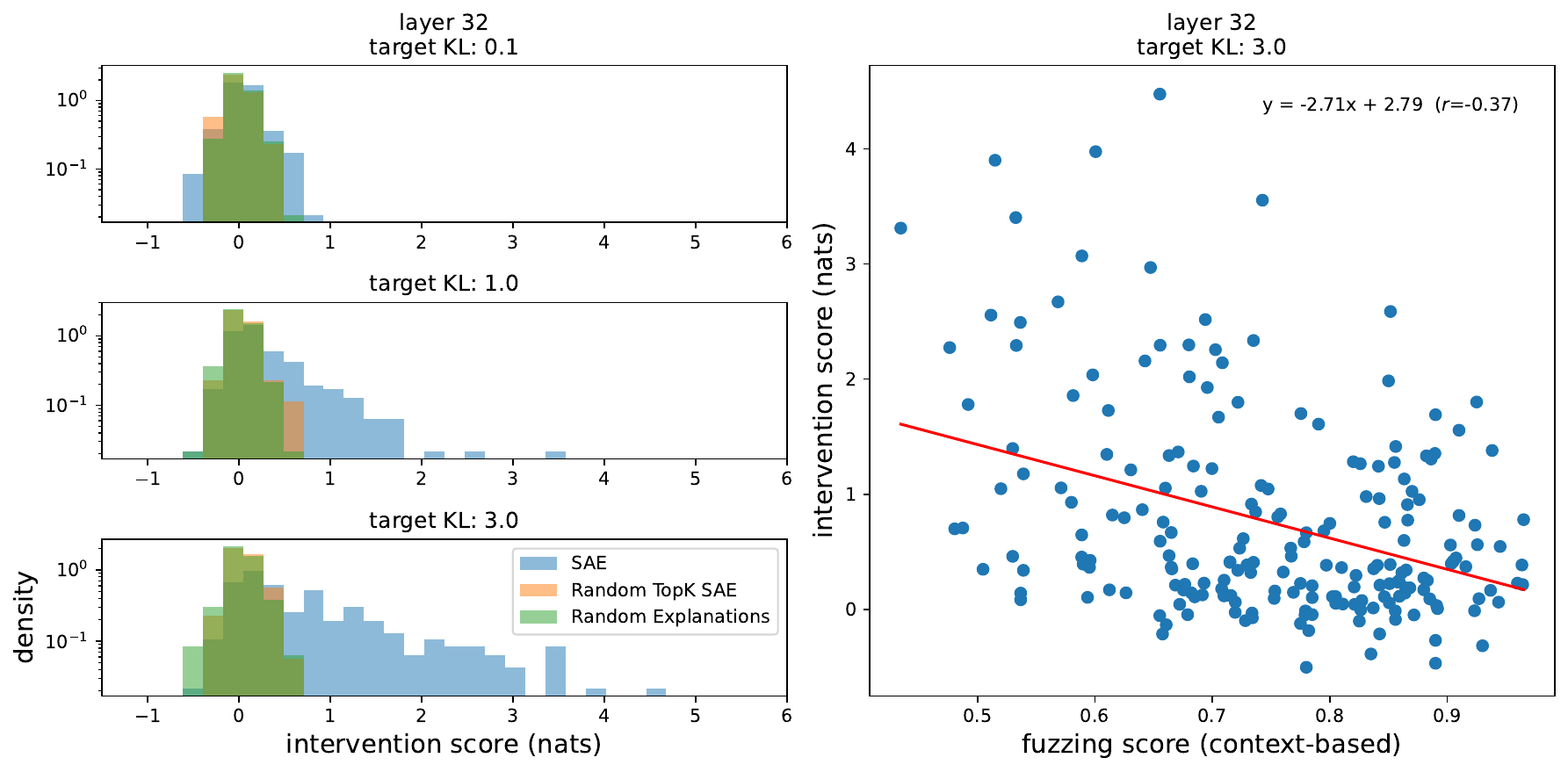}
    \vskip -0.2cm
    \caption{\textbf{Intervention scores.} Here we present intervention scores (Sec~\ref{sec:output-scoring}) for SAE features in Gemma 2 9B at layer 32. \textbf{Left:} SAE features are more interpretable than random features, especially when intervening more strongly. Our explainer also produces interpretations that are scored higher than random interpretations. \textbf{Right:} Many features that would normally be uninterpreted when using context-based automatic interpretability are interpretable in terms of their effects on output.} 
    \label{fig:intervention-hists-and-scatter}
\end{figure*}

In Figure~\ref{fig:intervention-hists-and-scatter}, we see that there is a slight negative correlation between the fuzzing score and the intervention score, showing that on average, features with low fuzzing score can be better explained using their downstream effects than features with high fuzzing score. We also find that the distribution of intervention scores of interpretation on a trained SAE is significantly different from the distribution of an interpretation of a randomly initialized SAE or a real SAE with randomly assigned interpretations, supporting the validity of this scoring method.

\subsection{Comparing interpretation methods}

We use a set of 500+ features as a testbed to measure the effects of design choices and hyperparameters on interpretation quality. Each feature is scored using 100 activating and 100 non-activating examples. The activating examples are chosen via stratified sampling such that there are always 10 examples from each of the 10 deciles of the activation distribution. We evaluate interpretation quality using fuzzing, detection and embedding scores, as those were both quick to compute and easy to interpret. We expect this mix of scores reflect the extent to which the proposed interpretation is valid over both activating and non-activating examples.

We find that the interpretations generated by considering only the activating contexts are significantly different from those generated by selecting activating contexts from the whole distribution. Interpretations based on top examples have higher specificity, but lower sensitivity. In fact, we observe that the sensitivity depends on how the degree of activation of each example shown, and that this sensitivity decreases faster for interpretation based on top examples. Interpretations generated from top activating examples are more concise and specific interpretations but fail to capture the whole distribution, while explanations based on examples sampled from the whole distribution lead to interpretations that are too broad, see Appendix \ref{app:examples_contexts}. This effect would not be seen if the scoring were done on just the most activating examples, underscoring a problem with current auto-interpretability evaluations, which produce interpretations using top activating examples and evaluate them on a small subset of the activation distribution.

Increasing the size of the explainer model increases the scores of the interpretations, but we don't find the explanations generated by Claude Sonnet 3.5 to have much higher scores than those generated by Llama 3.1 70b, see \ref{tab:model_size}, and both are similar to those generated by a human. Not surprisingly, we also see that even for simpler scores like fuzzing and detection, using a smaller scorer model leads to lower scores. It is possible the explanations generated by Claude would be better than those generated by Llama 3.1 70b had we optimized the prompting techniques for that model.

Showing the explainer model a larger number of examples leads to slightly higher scores (Table~\ref{tab:number_examples}). We find it doesn't matter much whether we use the same dataset used to train the SAEs (Table~\ref{tab:dataset}) or whether we slightly change the size of the contexts shown (Table~\ref{tab:context}). Using COT, at least when generating interpretations using Llama 70b, does not significantly increase their quality (Table~\ref{tab:cot}) while significantly increasing the compute and time required to generate them. For this reason, we have not used it for our main experiments. 

We find that SAEs with more features have higher scores, and scores that are significantly higher than those of neurons. Neurons are more interpretable if made sparser by only considering the top $k$ most activated neurons on a given token, but still significantly underperform SAEs in our tests (Table~\ref{tab:feature_size}). The location of the SAE matters; residual stream SAEs have slightly better scores than ones trained on MLP outputs (Table~\ref{tab:feature_size}). We also observe that earlier layers have lower overall scores, but that the scores across model depth remain constant after the first few layers (Fig~\ref{fig:layer_scores}).

\section{Conclusion}

Explaining the features of SAEs trained on cutting-edge LLMs is a computationally demanding task, requiring scalable methods to both generate interpretations and assess their quality. We addressed issues with the conventional simulation-based scoring and introduced five new scoring techniques, each with distinct strengths and limitations. These methods allowed us to explore the ``prompt design space'' for generating effective interpretations and propose practical guidelines.

Additionally, although current scoring methods do not account for interpretation length, we believe shorter interpretations are generally more useful and will incorporate this in future metrics. Some scoring methods also require further refinement, particularly in selecting non-activating examples to improve evaluation. We also investigated a method to generate and score interpretations of features based on their downstream effect and shown that low score ``correlational'' scores could be due to the existence of ``output'' features.

Access to better, automatically generated interpretations could play a crucial role in areas like model steering, concept localization, and editing. We hope that our efficient scoring techniques will enable feedback loops to further enhance the quality of interpretations.

\section*{Data and Code Availability}
Together with this manuscript, we release the \href{https://github.com/EleutherAI/delphi}{delphi}
library and accompanying scripts to guide the reproduction of these results.

In collaboration with Neuronpedia we also release the explanations for the Gemma 2 9b \href{https://www.neuronpedia.org/gemma-2-9b/3-gemmascope-res-16k-l0_32plus/}{16k residual stream} SAEs, the \href{https://www.neuronpedia.org/gemma-2-9b/24-gemmascope-mlp-16k-l0_32plus/}{16k MLP} SAEs, the \href{https://www.neuronpedia.org/gemma-2-9b/gemmascope-res-131k-l0_32plus/}{131k residual stream} SAEs and the Llama 3.1 8b \href{https://www.neuronpedia.org/llama3.1-8b/eleuther_gp-res-262k }{262k residual stream and} \href{https://www.neuronpedia.org/llama3.1-8b/eleuther_gp-mlp-262k}{MLP} SAEs. Explanations can be directly downloaded \href{https://huggingface.co/datasets/EleutherAI/auto_interp_explanations/}{here}.

\section*{Acknowledgements}
We would like to acknowledge Lucia Quirke and David Johnston for the work reviewing and discussing the manuscript. We would like to acknowledge Jacob Drori's role in the initial development of new scoring techniques. We would also like to thank Joseph Bloom, Sam Marks, Can Rager, Jannik Brinkmann and Sam Marks, Neel Nanda, Sarah Schwettmann and Jacob Steinhardt for their discussions and suggestions in an earlier version of the work. We are thankful to \href{https://www.openphilanthropy.org/grants/eleuther-ai-interpretability-research/}{Open Philanthropy for funding this work}.
We are grateful to CoreWeave for providing the compute resources.

\section*{Impact Statement}

This paper presents work whose goal is to advance the field of Mechanistic Interpretability. There are many potential societal consequences of our work, none which we feel must be specifically highlighted here.


\bibliography{example_paper}
\bibliographystyle{icml2025}

\newpage
\appendix
\onecolumn
\renewcommand{\thefigure}{A\arabic{figure}}

\setcounter{figure}{0}

\section{SAE interpretations of a random sentence}

In this section, we show some non cherry-picked examples of feature interpretations on a sentence.

\begin{figure}[h]
    \centering
    \includegraphics[width=1\linewidth]{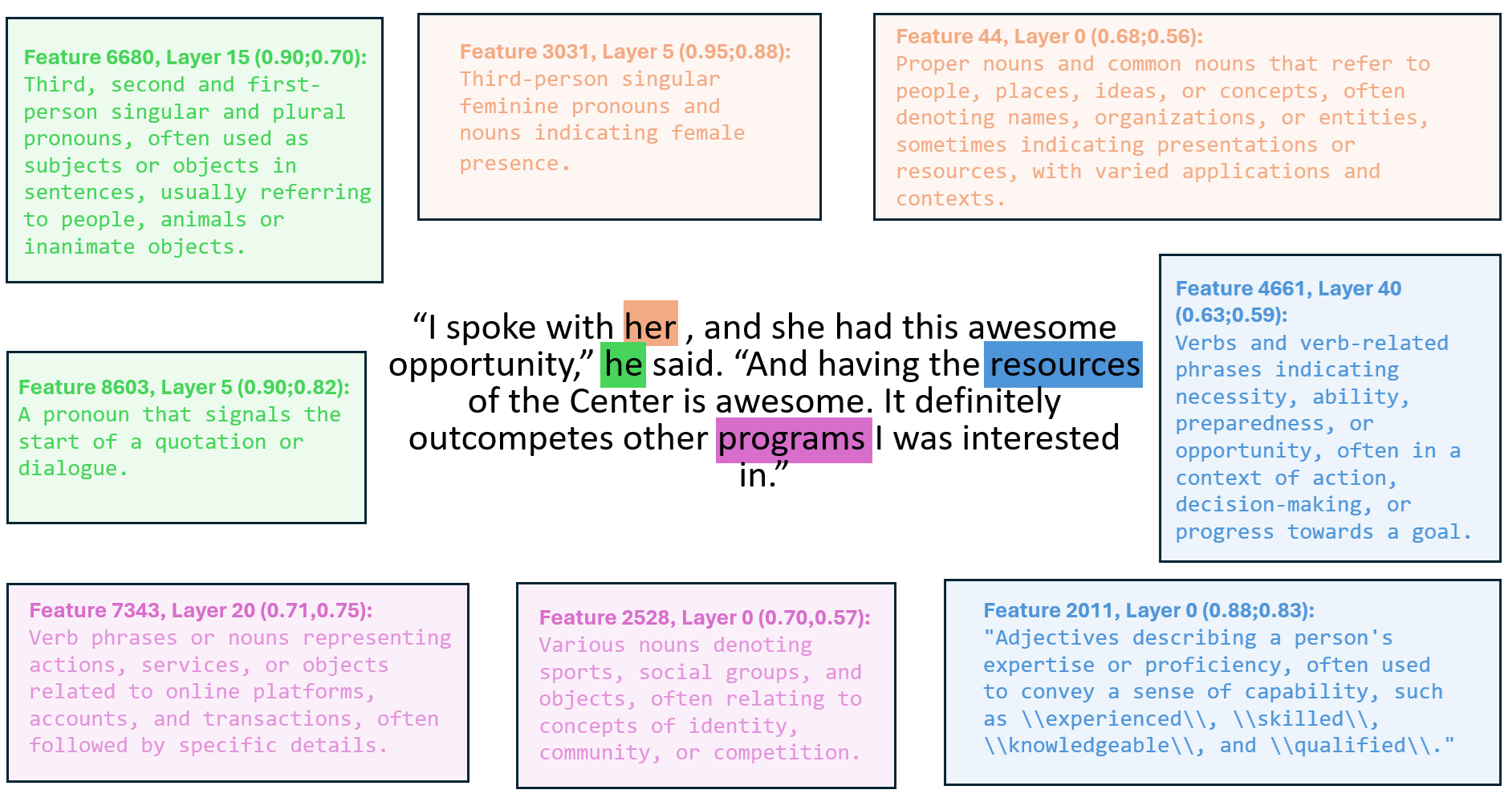}
    \caption{\textbf{SAE features interpretations for a random sentence}. To visualize the feature interpretations produced, we select a sentence taken from the RPJv2 dataset. We selected 4 tokens in different positions in that sentence and filter for features that are active in different layers. Then we randomly select active features and their corresponding interpretations to display. We display the detection and fuzzing scores of each interpretation, which indicate how well it explains other examples in the dataset (see Section 3 for details on these scores). The features selected had high activation, but were not cherry-picked based on interpretations or scores.}
    \label{fig:5Mfeatures}
\end{figure}

\section{Explainer Prompt}
\label{app:explainer-prompt}

The system prompt if not using Chain of Thought (COT):
\begin{mdframed}
\begin{verbatim}
You are a meticulous AI researcher conducting an important 
investigation into patterns found in language. Your task is to
analyze text and provide an interpretation that thoroughly 
encapsulates possible patterns found in it.
Guidelines:

You will be given a list of text examples on which special words
are selected and between delimiters like << this >>. 
If a sequence of consecutive tokens all are important, 
the entire sequence of tokens will be contained between 
delimiters <<just like this>>. How important each token is for
the behavior is listed after each example in parentheses.

- Try to produce a concise final description. Simply describe
the text features that are common in the examples, and what
patterns you found.

- If the examples are uninformative, you don't need to mention
them. Don't focus on giving examples of important tokens,
but try to summarize the patterns found in the examples.

- Do not mention the marker tokens ($<<$ $>>$) 
in your interpretation.

- Do not make lists of possible interpretations.
Keep your interpretations short and concise.

- The last line of your response must be the formatted 
interpretation, using [interpretation]:
\end{verbatim}
\end{mdframed}
We add to the previous prompt the following if we want to do COT:
\begin{mdframed}
\begin{verbatim}
To better find the interpretation for the language patterns,
go through the following stages:

1.Find the special words that are selected in the examples
and list a couple of them. Search for patterns in these words,
if there are any. Don't list more than 5 words.

2. Write down general shared features of the text examples.
This could be related to the full sentence or to the words 
surrounding the marked words.

3. Formulate a hypothesis and write down the final interpretation
using [interpretation]:.
\end{verbatim}
\end{mdframed}
One of the few shot examples of how examples are displayed to the model.
\begin{mdframed}
\begin{verbatim}
Example 1:  and he was <<over the moon>> to find

Activations: (``over", 5), (`` the", 6), (`` moon", 9)

Example 2:  we'll be laughing <<till the cows come home>>! Pro

Activations: (``till", 5), (`` the", 5),
(`` cows", 8), (`` come", 8),
(`` home", 8)

Example 3:  thought Scotland was boring, but really there's
more <<than meets the eye>>! I'd

Activations: (``than", 5), (`` meets", 7), (`` the", 6), 
(`` eye", 8)
\end{verbatim}
\end{mdframed}
If COT is used, an explicit example of using COT is demonstrated.
\begin{mdframed}
\begin{verbatim}
ACTIVATING TOKENS: ``over the moon", ``than meets the eye".
SURROUNDING TOKENS: No interesting patterns.

Step 1.
- The activating tokens are all parts of common idioms.
- The surrounding tokens have nothing in common.

Step 2.
- The examples contain common idioms.
- In some examples, the activating tokens are followed 
by an exclamation mark.

Step 3.
- The activation values are the highest for the more common
idioms in examples 1 and 3.

Let me think carefully. Did I miss any patterns in the text
examples? Are there any more linguistic similarities?

- Yes, I missed one: The text examples all convey positive
sentiment.
\end{verbatim}
\end{mdframed}
Afterwards an interpretation is added to the example
\begin{mdframed}
\begin{verbatim}
[interpretation]: Common idioms in text conveying positive
sentiment.
\end{verbatim}
\end{mdframed}

\section{Examples of activating contexts and interpretations}
\label{app:examples_contexts}
\begin{figure}[h!]
    \centering
    \includegraphics[width=0.7\linewidth]{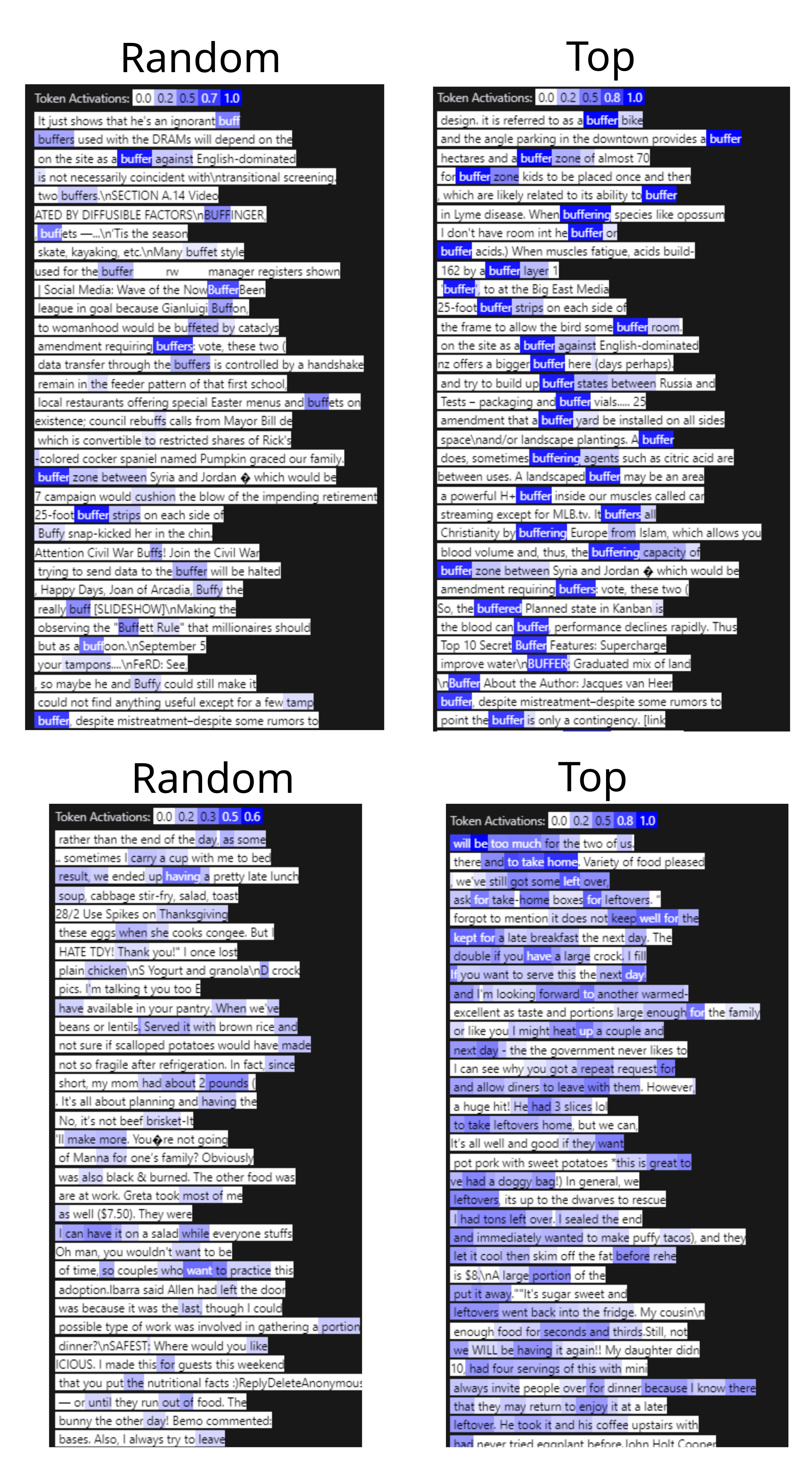}
    \caption{Activating contexts of feature 209 (top) and 293 (bottom), from layers 8 and 32, respectively, of the 131k feature SAE trained on the residual stream of Gemma 2 9b. The shown examples are similar to the ones given to the explainer model to come up with interpretations.}
    \label{fig:top_vs_random}
\end{figure}

As discussed in the main text, the interpretations found for a given feature can be very different depending on the way to sample the activating contexts shown to the explainer model, see Figure \ref{fig:top_vs_random}. This has its advantages and disadvantages.

When using the top activating contexts, the explainer model normally gives an interpretation that is more narrow - ``The concept of a buffer, referring to something that separates, shields, or protects one thing from another, often used in various contexts such as physical barriers, chemical reactions, or digital data processing" - instead of one that captures the full distribution - ``Words or phrases associated with concepts of spatial or temporal separation (buffers, zones, or cushions) or colloquialisms (buff, buffs, or Buffy, referring to a popular TV show or enthusiast)". Here the narrower interpretations resulted in lower scores: the interpretation generated from examples sampled from the whole distribution scored 0.95 accuracy in fuzzing and 0.93 accuracy in detection, while the interpretation generated from top examples only achieves 0.8 accuracy in fuzzing and 0.74 accuracy in detection.

On the other hand, if we look at the second example, sampling from the whole distribution examples may sometimes confuse the explainer model - ``Tokens often precede or succeed prepositions, articles, and words that signal possession or quantity, frequently indicating a relationship between objects or actions. of feature", which might not see the pattern that is more clear in the top activating examples -``Descriptions of food or events where food is involved, often mentioning leftovers, and sometimes mentioning the act of eating, serving, or storing food, as well as the amount of food, or the pleasure or satisfaction derived from it." Here we see very poor scores on the interpretation from randomly sampled examples, 0.54 accuracy in detection and 0.57 accuracy in fuzzing, while the interpretation generated from the top examples has 0.89 accuracy both in detection and fuzzing,

\section{Different scoring methods}
\label{app:scoring_details}
\subsection{Detection details}
\label{app:detection_details}
The prompt used for detection scoring is the following:
\begin{mdframed}
\begin{verbatim}
You are an intelligent and meticulous linguistics researcher.

You will be given a certain feature of text, such as 
``male pronouns" or ``text with negative sentiment".

You will then be given several text examples. Your task 
is to determine which examples possess the feature.

For each example in turn, return 1 if the sentence is 
correctly labeled or 0 if the tokens are mislabeled. You must
return your response in a valid Python list. Do not return
anything else besides a Python list.
\end{verbatim}
\end{mdframed}
Together with the prompt, there are several few shot examples like the following:
\begin{mdframed}
\begin{verbatim}
<user_prompt>

feature interpretation: Words related to American football
positions, specifically the tight end position.

Text examples:

Example 0:Getty Images Patriots tight end Rob Gronkowski
had his boss

Example 1: names of months used in The Lord of the Rings:
the

Example 2: Media Day 2015 LSU defensive end Isaiah Washington
(94) speaks to the

Example 3: shown, is generally not eligible for ads. For 
example, videos about recent tragedies,

Example 4: line, with the left side namely tackle Byron
Bell at tackle and guard Amini

<assistant_response>

[1,0,0,0,1]    
\end{verbatim}
\end{mdframed}
In this example the $<$user\_prompt$>$ and the $<$assistant\_response$>$ tags are substituted with the correct instruct format used by the scorer model. In detection scoring, 5 shuffled examples are shown to the model at the same time.

\subsection{Fuzzing details}
\label{app:fuzzing_details}
The prompt used for fuzzing scoring is the following:
\begin{mdframed}
\begin{verbatim}
    
You are an intelligent and meticulous linguistics researcher.

You will be given a certain feature of text, such as ``male 
pronouns" or ``text with negative sentiment". You will 
be given a few examples of text that contain this feature.
Portions of the sentence which strongly represent this 
feature are between tokens << and >>.

Some examples might be mislabeled. Your task is to determine
if every single token within << and >> is correctly
labeled. Consider that all provided examples could be correct,
none of the examples could be correct, or a mix. An example
is only correct if every marked token is representative
of the feature

For each example in turn, return 1 if the sentence is correctly
labeled or 0 if the tokens are mislabeled. You must return
your response in a valid Python list. Do not return anything
else besides a Python list.
\end{verbatim}
\end{mdframed}

Followed by few-shot examples of the kind:
\begin{mdframed}
\begin{verbatim}
<user_prompt>

feature interpretation: Words related to American football
positions, specifically the tight end position.

Text examples:

Example 0:Getty Images Patriots<< tight end>> Rob Gronkowski
had his boss

Example 1: posted You should know this<< about>> offensive
line coaches: they are large, demanding<< men>>

Example 2: Media Day 2015 LSU<< defensive>> end Isaiah
Washington (94) speaks<< to the>>

Example 3:<< running backs>>,`` he said. .. Defensive
<< end>>
Carroll Phillips is improving and his injury is

Example 4:<< line>>, with the left side namely<< tackle>>
Byron Bell at<< tackle>> and<< guard>> Amini

<assistant_response>

[1,0,0,1,1]
\end{verbatim}
\end{mdframed}
In this example, the $<$user\_prompt$>$ and the $<$assistant\_response$>$ tags are substituted with the correct instruct format used by the scorer model. In fuzzing scoring, 5 shuffled examples are shown to the model at the same time.

\subsection{Surprisal details}
\label{app:surprisal_details}
In surprisal scoring, the cross-entropy loss of the scorer model is computed over the tokens of an example. For each interpretation, this loss is computed with the interpretation and with a default interpretation - ``Various unrelated sentences,'' where the examples can either be activating context or non-activating contexts. In our first approach, Llama 3.1 70b base was used. The prompt starts with few shot examples like:
\begin{mdframed}
\begin{verbatim}
The following is a description of a certain feature of text 
and a list of examples that contain the feature.

Description: 

References to the Antichrist, the Apocalypse and conspiracy
theories related to those topics. 

Sentences: 

 `` by which he distinguishes Antichrist is, that he would
 rob God of his honour and take it to himself, he gives
 the leading feature which we ought  "
 
 ``3 begins. And the rise of Antichrist. Get ready with   " 
 
 `` would be destroyed. The worlds economy would likely
 collapse as a result and could usher in a one world government
 movement. I wrote a small 6 page  " 
 
Description:  
 
Sentences containing digits forming a four-digit year.
 
Sentences: 
 
 `` 20, 2013 at 7:41 pm Martin Smith  "
 
 `` of 2012. In other words, Italy's   "
 
 ``end 2012 levels). In the first quarter of 2013, we expect
 revenue to be up slightly from the fourth quarter  "
 
Description: 
 
Text related to banking and financial institutions.

Sentences: 
 
 ``: He is on the Board of Directors with the Lumbee Bank   "
 
 `` refurbishing the Bank’s branches. BIP reached 400
 thousand users in one year The use of BIP has already
 doubled The  "
 
 `` the Federal Deposit Insurance Corp.  "
 
Description:  
 
Occurrences of the word 'The' at the beginning of sentence.
 
Sentences:  
 
 ``The Smoking Tire hits the canyons with one of the fastest
 Audi's on the road  "
 
 ``The Chairman of the ABI  "
 
 ``The administrative center is the town of Koch.  "
\end{verbatim}
\end{mdframed}

With the pairs of losses with the interpretation and with the default interpretation, it is possible to compute the decrease in loss caused by having access to the interpretation. It is expected that in activating contexts this difference will be greater than in non-activating examples, and the score of the interpretation is the AUC computed using this loss as a proxy for activating and non-activating labels.

\subsection{Embedding details}
\label{app:embedding_details}
\begin{figure}[h]
    \centering
    \includegraphics[width=1\linewidth]{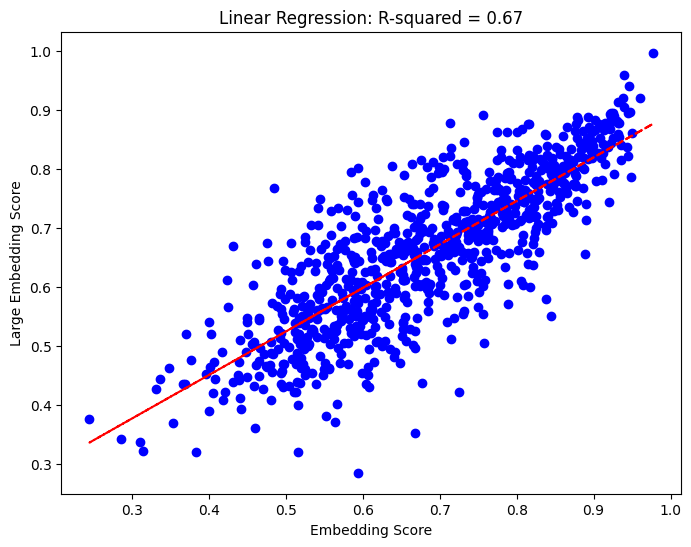}
    \caption{\textbf{Comparison between the scores given by a small embedding model and a larger one}}
    \label{fig:embedding}
\end{figure}

For embedding scoring, we use a \href{https://huggingface.co/dunzhang/stella_en_400M_v5}{small 400M}  parameter model. We chose a small performant model on MTEB \citep{muennighoff2022mteb} because we found similar scores when using a larger \href{https://huggingface.co/nvidia/NV-Embed-v2}{7B parameter model}, see fig \ref{fig:embedding}, as this size allowed us to do increase the number of examples used in scoring.

A set of activating and randomly selected non-activating examples are embedded using the scorer model. Then a ``query'' instruction is embedded: 
\begin{mdframed}
\begin{verbatim}    
Instruct: Retrieve sentences that could be related to 
the interpretation. Query:  \{interpretation\}
\end{verbatim}
\end{mdframed}

The cosine-similarity between the instruction embedding and the examples is computed and is used as a proxy for activating and non-activating labels when computing the AUC, which is the score of that interpretation.

\subsection{Adversarial examples}

Most features are active in less than 0.1\% of the full dataset used to collect the activations, making random non-activating contexts very diverse. Randomly sampling non-activating examples cannot be used to determine whether a feature fires in a token in a specific context or on that token in general, as it is unlikely that that token randomly occurs in each non-activating example. As SAEs are scaled and features become sparser and more specific, techniques that overly rely on activating contexts will have more imprecise results \cite{gao2024scaling}.

Motivated by the phenomenon of feature splitting, we could use ``similar'' features to test whether interpretations are precise enough to distinguish between similar contexts. A potential approach is using cosine similarity between decoder directions of features to find counterexamples for an interpretation. Some works \citep{eleuther_auto-2024,goodfire-2024} have shown that a significant fraction of current feature interpretations can't be used to distinguish between features with high similarity.

\subsection{Correlation between scores}
\label{app:score_correlation}
We compute the correlation between these different scores and simulation 800 different feature interpretations, of the 131k feature SAE trained on the residual stream of Gemma 2 9b spread across  4 different layers. We use 100 activating example, 10 from each of 10 different activating quantiles, and 100 non-activating examples to compute these scores and use Llama 3.1 70b Instruct as the scorer model for simulation, fuzzing and detections, while we use Llama 3.1 70b Base for surprisal and \href{dunzhang/stella_en_400M_v5}{Stella-400M} as the embedding model.

Interestingly, we also find that the correlation between simulation scores and other scores also increases when the simulation scorer is Claude Sonnet 3.5. These estimations are on a smaller number of features(100), due to the high cost of simulation using Claude Sonnet, but on this set, the Spearman correlation between fuzzing and simulation increases from 0.69 to 0.75, between detection and simulation increases from 0.31 to 0.38, between embedding and simulation from 0.14 to 0.21 and from surprisal to simulation from 0.02 to 0.10. The baseline numbers for the correlation of simulation scores are lower on this set, so we expect that had we done Claude scores on a larger set of explanations, the correlations between the scores would be higher.

\renewcommand{\thetable}{A\arabic{table}}

\setcounter{table}{0}

\begin{table}[h]
\caption{Spearman correlation, computed over scores of 800 different feature interpretations. }
\label{tab:spearman}
\centering
\begin{tabular}{lccccc}
\toprule
           & Fuzzing & Detection & Simulation & Embedding & Surprisal \\
\midrule
Fuzzing    & 1       & 0.73      & 0.75       & 0.41      & 0.30      \\
Detection  &         & 1         & 0.44       & 0.71      & 0.62      \\
Simulation &         &           & 1          & 0.28      & 0.15      \\
Embedding  &         &           &            & 1         & 0.79      \\
Surprisal  &         &           &            &           & 1         \\
\bottomrule
\end{tabular}
\end{table}

\begin{table}[t]
\caption{Pearson correlation computed over 800 different feature scores}
\label{tab:pearson}
\centering
\begin{tabular}{lcccccc}
\toprule
           & Fuzzing & Detection & Simulation & Embedding & Surprisal \\
\midrule
Fuzzing    & 1       & 0.74      & 0.74       & 0.42      & 0.32     \\
Detection  &         & 1         & 0.46       & 0.70      & 0.62    \\
Simulation &         &           & 1          & 0.30      & 0.14    \\
Embedding  &         &           &            & 1         & 0.79   \\
Surprisal  &         &           &            &           & 1       \\
\bottomrule
\end{tabular}
\end{table}

\begin{figure}[h]
    \centering
    \includegraphics[width=1\linewidth]{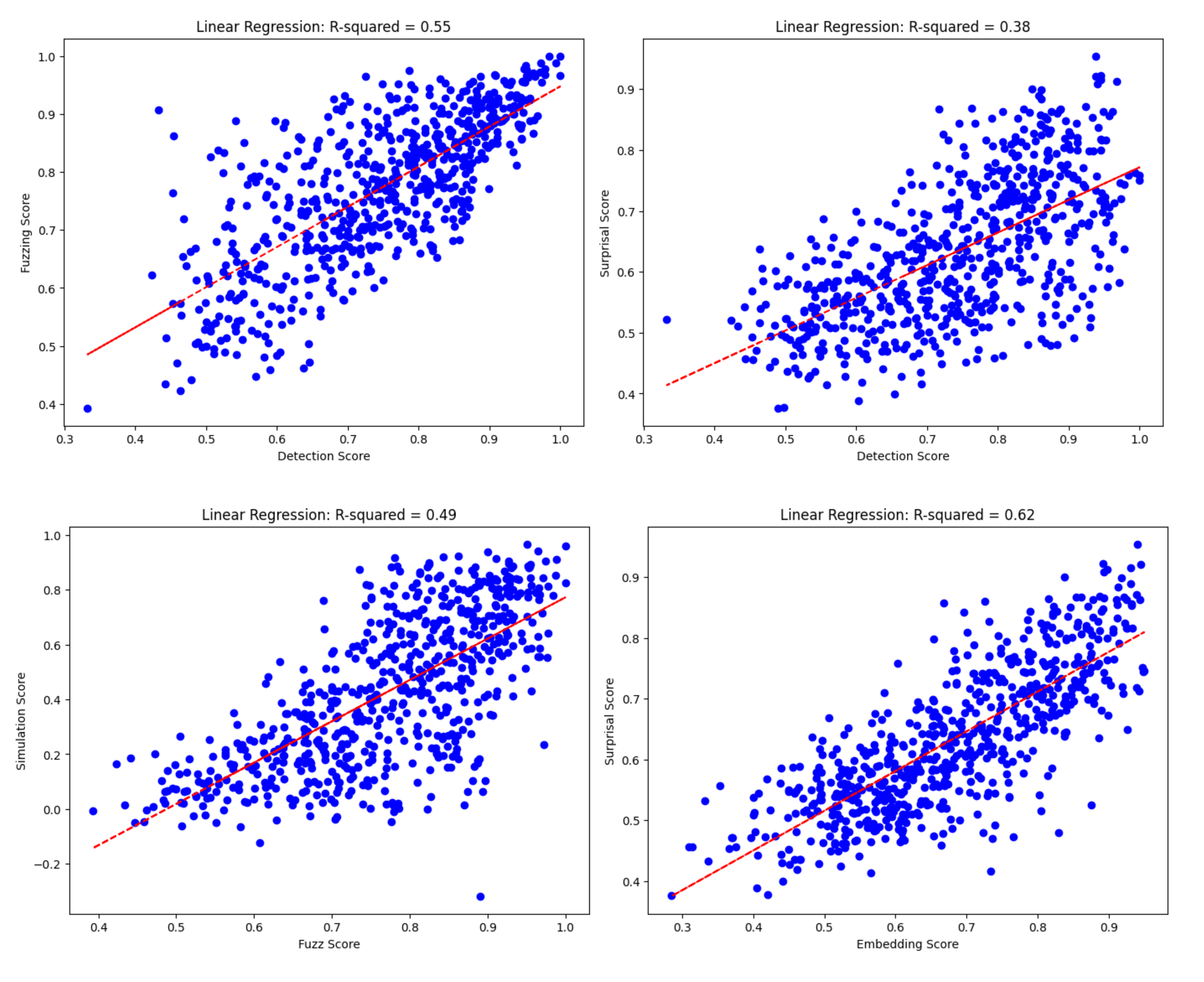}
    \caption{\textbf{Scatter plots with different combinations of scores}}
    \label{fig:correlation}
\end{figure}

\subsection{Disagreement between scores}

In this section, we will discuss how different types of scores might evaluate explanations differently. In particular, we will focus on an interpretation with fuzzing score but low detection score, high fuzzing score but low simulation score, high simulation score but low embedding score and high embedding score and low simulation score. 

For instance, feature 2 of layer 8 of the 131k feature SAE model trained on the residual stream of Gemma 2 9b has a fuzzing score of 0.9 but a detection score of only 0.43. The interpretation given by our pipeline is "Verbs that link a subject to additional information, often in a formal or descriptive tone." Most of the activations of this feature are on sentences like "This $<$is needed$>$ because Magento (...)" or "For instance, this$<$ is$>$ technically correct syntax", where $< >$ represent active tokens. In both these examples, the model correctly identifies the highlighted tokens as active during fuzzing, but incorrectly identifies the context as non-active during detection. On the other hand, detection incorrectly identifies sentences like "pay for things that would prevent larger issues down the road is better in the long run." or "Neuroscientist Jack Gallant calls the research a technologic tour de force and says the ultimate decoder would provide vivid" as active. This explanation is too vague on the activating context, leading to a low detection score, but specific enough in the types of tokens that are active, leading to a high fuzz score.

Feature 281 of the same layer has a fuzzing score of 0.97, but a simulation score of only 0.19. Its interpretation is "URLs or hyperlinks containing query parameters, indicating a request for specific data or actions on a web page." The simulation score is lower for than the fuzzing score because the model has to decide which parts of the URL the feature is active on, while the fuzzing scorer either is shown links that are highlighted which it will say are active or non-links that are highlighted and are, most likely, not active.

Feature 8 of layer 24 has a simulation score of 0.73, but an embedding score of only 0.43. Its interpretation is "Prevalence of logical operators and conjunctions in text, including simple addition and conjunction in various contexts, as well as indicators of contrasting or additional information, often used for comparison or to provide supplementary details". While doing simulation, the model correctly identifies that in the sentence "the cost of the unit itself, plus installation and construction costs.", the token "plus" is active but all the others aren't. The same is true for sentences like "Plus my potato bread I made on the ANZAC Day" and "Cliff Chiang on reintroducing Orion in Wonder Woman! Plus interviews". On the other hand, these sentences don't have high embedding similarity with the explanation.

Feature 10 of layer 16 has the opposite situation happening, with an embedding score of 0.74 and a simulation score of only 0.04. Its interpretation is "Verbs, prepositions, and adverbs that connect clauses or indicate direction, movement, or progression, often in a sequence of actions or events". In sentences like "right, three months <to close down>. There was dead silence when the message was read. Everybody waited for Mr. Smith to speak. Mr. Gingham" the simulator model incorrectly identifies " waited" and the second " to" as active, and misses the real activation in "to close down". The same happens in "This is <going> to <be> one <swinging> party. Especially if the guests survive Uncle Wilhelm. I love weddings. I love going to them, I love being in", where the simulator model identified "love being in" as well as "love going to" as being active. The embedding model correctly identified that most activating sentences have the mention of motion. 

With these examples, we hope to demonstrate that these generated explanations are not perfect, but that we have an easier understanding of their flaws by looking at which cases different scores disagree on.

\section{SAE models used}

Throughout this work, we used different SAEs trained on Gemma. The 16k feature ones trained on the MLP have the following average L0 norms per layer:
\begin{mdframed}
0:50, 1:56,
2:33, 3:55,
4:66, 5:46,
6:46, 7:47,
8:55, 9:40,
10:49, 11:34,
12:42, 13:40,
14:41, 15:45,
16:37, 17:41,
18:36, 19:38,
20:41, 21:34,
22:34, 23:73,
24:32, 25:72,
26:57, 27:52,
28:50, 29:49,
30:51, 31:43,
32:44, 33:48,
34:47, 35:46,
36:47, 37:53,
38:45, 39:43,
40:37, 41:58
\end{mdframed}
The 16k feature ones trained on the residual stream have the following average L0 norms:
\begin{mdframed}
0:35, 1:69,
2:67, 3:37,
4:37, 5:37,
6:47, 7:46,
8:51, 9:51,
10:57, 11:32,
12:33, 13:34,
14:35, 15:34,
16:39, 17:38,
18:37, 19:35,
20: 36, 21:36,
22: 35, 23: 35,       
24: 34, 25: 34,
26: 35, 27:36,
28: 37, 29:38,
30:37, 31:35,
32: 34, 33:34,
34:34, 35:34,
36:34, 37:34,
38:34, 39:34,
40:32, 41:52
\end{mdframed}
The 131k feature ones trained in the residual steam have the following average L0 norms:
\begin{mdframed}
0:30,
 1:33, 2:36, 3:46,
 4:51, 5:51,
 6:66, 7:38,
 8:41, 9:42,
 10:47, 11:49,
 12:52, 13:30,
 14:56, 15:55,
 16:35, 17:35,
 18:34, 19:32, 
 20:34, 21:33,
 22:32, 23:32,
24: 55, 25:54,
26:32, 27:33,
28: 32, 29:33,
30:32, 31:52,
32: 51, 33:51,
34:51, 35:51,
36: 51, 37:53,
38:53, 39:54,
40: 49, 41:45
\end{mdframed}

We also used SAEs with 65k features, trained on the residual stream and the MLP of Llama 8b. 

\section{Factors that influence the explainability}
\label{app:explanability_factors}
\subsection{Dependence on dataset}
Even though a significant portion of SAE features are less active when using RPJv2 instead of the Pile, we find that the features that are active are generally interpretable to the same degree. These evaluations were done using the 131k feature SAE trained on the residual stream of Gemma 2 9b. The scorer and the explainer model where Llama 3.1b 70b instruct, quantized to 4bit.
\begin{table}
\caption{Impact of training dataset on Fuzzing and Detection performance. Numbers shown are the median score and the interquartile (25\%-75\%) range.}
\label{tab:dataset}
\centering
\begin{tabular}{lccc}
\toprule
Experiment & Fuzzing & Detection & Embedding\\
\midrule
Random interpretation & 0.51 (0.45--0.57) & 0.51 (0.45--0.58) & 0.51 (0.44--0.57)\\
Randomly initialized Topk SAE & 0.55 (0.50--0.60) & 0.54 (0.50--0.59) & -- \\
\addlinespace
RPJ-v2 & 0.76 (0.67--0.86) & 0.74 (0.63--0.85) & 0.67 (0.57--0.80) \\
Pile & \textbf{0.76 (0.67--0.86)} & \textbf{0.76 (0.67--0.85)} & \textbf{0.69 (0.57--0.80)}\\
\bottomrule
\end{tabular}
\end{table}
\subsubsection{Dependence on chain of thought and activation information}
We find that COT slightly increases the scores of the interpretations found, and that it significantly slows down the rate at which one can produce interpretations. Giving the explainer model, the activations associated with each token seem to slightly increase the scores of the interpretations generated. These evaluations were done using the 131k feature SAE trained on the residual stream of Gemma 2 9b. The scorer and the explainer model where Llama 3.1b 70b instruct, quantized to 4bit.

\begin{table}
\caption{Impact of prompt content on Fuzzing and Detection performance. Numbers shown are the median score and the interquartile (25\%-75\%) range.}
\label{tab:cot}
\centering
\begin{tabular}{lccc}
\toprule
Experiment & Fuzzing & Detection & Embedding \\
\midrule
Random interpretation & 0.51 (0.45--0.57) & 0.51 (0.45--0.58) & 0.51 (0.44--0.57)\\
Randomly initialized Topk SAE & 0.55 (0.50--0.60) & 0.54 (0.50--0.59) & -- \\
\addlinespace
Activations in prompt & \textbf{0.76 (0.67--0.86)} & \textbf{0.74 (0.63--0.85)} & \textbf{0.68 (0.57--0.80)} \\
No activations in prompt & 0.75 (0.65--0.86) & 0.73 (0.60--0.84) & 0.68 (0.58--0.79) \\
COT in prompt & 0.76 (0.67--0.86) & 0.73 (0.61--0.85) & 0.65 (0.55--0.75)\\
\bottomrule
\end{tabular}
\end{table}

\subsection{Dependence on context length}

The context length of the shown examples did not significantly change the scores obtained for the interpretations. These evaluations were done using the 131k feature SAE trained on the residual stream of Gemma 2 9b. The scorer and the explainer model were Llama 3.1b 70b instruct, quantized to 4 bits.

\begin{table}
\caption{Impact of context length on Fuzzing and Detection performance. Numbers shown are the median score and the interquartile (25\%-75\%) range.}
\label{tab:context}
\centering
\begin{tabular}{lccc}
\toprule
Experiment & Fuzzing & Detection & Embedding \\
\midrule
Random interpretation & 0.51 (0.45--0.57) & 0.51 (0.45--0.58) & 0.51 (0.44--0.57)\\
Randomly initialized Topk SAE & 0.55 (0.50--0.60) & 0.54 (0.50--0.59) & -- \\
\addlinespace
16 context & 0.75 (0.65--0.86) & 0.74 (0.62--0.85) & 0.70 (0.59--0.81)\\
32 context & \textbf{0.76 (0.67--0.86)} & \textbf{0.74 (0.63--0.85)} & 0.67 (0.57--0.80) \\
64 context & 0.74 (0.64--0.64) & 0.70 (0.57--0.81) & 0.65 (0.54--0.78) \\
\bottomrule
\end{tabular}
\end{table}

\subsection{Dependence on the origin of examples}
Sampling the examples from the whole distribution of activations, or sampling a fixed number of examples over different activation quantiles, significantly improved the scores, compared with sampling only from the top examples. These evaluations were done using the 131k feature SAE trained on the residual stream of Gemma 2 9b. The scorer and the explainer model were Llama 3.1b 70b instruct, quantized to 4 bits.
\begin{table}
\caption{Impact of example sampling strategies on Fuzzing and Detection performance. Numbers shown are the median score and the interquartile (25\%-75\%) range.}
\label{tab:source_examples}
\centering
\begin{tabular}{lccc}
\toprule
Experiment & Fuzzing & Detection & Embedding \\
\midrule
Random interpretation & 0.51 (0.45--0.57) & 0.51 (0.45--0.58) & 0.51 (0.44--0.57)\\
Randomly initialized Topk SAE & 0.55 (0.50--0.60) & 0.54 (0.50--0.59) & -- \\
\addlinespace
Randomly sampled & 0.76 (0.68--0.86) & 0.74 (0.62--0.84) & 0.66 (0.56--0.78)  \\
Sampled from quantiles & \textbf{0.77 (0.69--0.87)} & \textbf{0.74 (0.64--0.85)} & 0.68 (0.57--0.80) \\
Sampled from top examples & 0.73 (0.64--0.83) & 0.72 (0.62--0.82) & \textbf{0.70 (0.58--0.80)} \\
\bottomrule
\end{tabular}
\end{table}

\subsection{Dependence on the number of examples}
The number of examples shown to the model seems to saturate, at least with the explainer model we used. These evaluations were done using the 131k feature SAE trained on the residual stream of Gemma 2 9b. The scorer and the explainer model where Llama 3.1b 70b instruct, quantized to 4bit.

\begin{table}
\caption{Impact of number of examples on Fuzzing and Detection performance. Numbers shown are the median score and the interquartile (25\%-75\%) range.}
\label{tab:number_examples}
\centering
\begin{tabular}{lccc}
\toprule
Experiment & Fuzzing & Detection & Embedding \\
\midrule
Random interpretation & 0.51 (0.45--0.57) & 0.51 (0.45--0.58) & 0.51 (0.44--0.57)\\
Randomly initialized Topk SAE & 0.55 (0.50--0.60) & 0.54 (0.50--0.59) & -- \\
\addlinespace
Shown 10 examples & 0.73 (0.62--0.85) & 0.71 (0.58--0.82) & 0.64 (0.54-0.74)\\
Shown 20 examples & 0.74 (0.64--0.85) & 0.72 (0.60--0.84) & 0.66 (0.54-0.76) \\
Shown 40 examples & \textbf{0.76 (0.67--0.86)} & \textbf{0.74 (0.63--0.85)} & \textbf{0.68 (0.57--0.80)} \\
Shown 60 examples & 0.75 (0.66--0.85) & 0.73 (0.62--0.84) & 0.68 (0.57--0.79) \\
\bottomrule
\end{tabular}
\end{table}

\subsection{Explainability across layers}

We find that, in the case of the 131k feature SAE trained on the residual stream, the earliest layers have lower scores than the later layers. These evaluations were done using the 131k feature SAE trained on the residual stream of Gemma 2 9b. The scorer and the explainer model where Llama 3.1b 70b instruct, quantized to 4bit.

\begin{figure}
    \centering
    \includegraphics[width=1\linewidth]{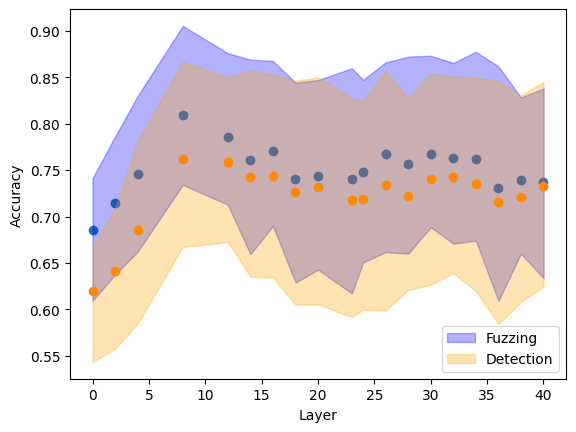}
    \caption{\textbf{Accuracy on fuzzing and detection scoring} Dots correspond to the median score over c.a. 300 feature interpretations, and colored region denotes the interquartile range.}
    \label{fig:layer_scores}
\end{figure}

\subsection{Dependence on size of explainer models}

Having a larger explainer model leads to better interpretations, but our results suggest that this benefit tends to saturate, as we find that the scores of explanations given by Claude or Llama 70b are very similar. We only did human interpretation of 100 features. Simulations scores are also only computed over 100 features. 

\begin{table}
\caption{Comparison of scores of interpretations given by different models. Numbers shown are the median score and the interquartile (25\%-75\%) range.}
\label{tab:explainer_size}
\centering
\begin{tabular}{lcccc}
\toprule
Experiment & Fuzzing & Detection & Embedding & Simulation \\
\midrule
Random interpretation & 0.51 (0.45--0.57) & 0.51 (0.45--0.58) & 0.51 (0.44--0.57) & -0.02 (-0.02--0.00) \\
\addlinespace
Claude & 0.75 (0.68--0.84) & \textbf{0.75 (0.65--0.85)} & 0.70 (0.58--0.81) & 0.30 (0.28--0.32)\\
Llama 70b  & \textbf{0.76 (0.67--0.86)} & 0.74 (0.63--0.85) & 0.67 (0.57--0.80) & 0.29 (0.28-0.32)\\
Llama 8b  & 0.70 (0.60--0.81) & 0.70 (0.59--0.81) & 0.64 (0.54--0.75) & 0.26 (0.23-0.30)\\
Human & 0.75 (0.66--0.85) & 0.74 (0.64--0.85) & \textbf{0.71 (0.62--0.81)} & \textbf{0.36 (0.32--0.39)}\\
\bottomrule
\end{tabular}
\end{table}

\subsection{Dependence on size of explainer models}

We find that both fuzzing, detection and simulation scoring are dependent on the size of the explainer model, that is, given the same explanations, larger models achieve higher scores on average. While Claude and Llama 80b have similar fuzzing and detection scores, simulation, being more complex, is better performed by Claude.

\begin{table}
\caption{Comparison of scorer models for fuzzing, detection and simulation. Numbers shown are the median score and the interquartile (25\%-75\%) range.}
\label{tab:model_size}
\centering
\begin{tabular}{lccc}
\toprule
Experiment & Fuzzing & Detection & Simulation \\
Claude & \textbf{0.76 (0.66--0.88)} & 0.72 (0.58--0.87) & 0.33 (0.29--0.36)\\
Llama 70b & 0.76 (0.67--0.86) & \textbf{0.74 (0.63--0.85)} & 0.29 (0.28--0.32) \\
Llama 8b  & 0.70 (0.60--0.81) & 0.70 (0.59--0.81) &  0.26 (0.25--0.31)\\
\bottomrule
\end{tabular}
\end{table}

\subsection{Dependence on SAE size and location}
We compare the interpretability of SAEs with different number of features, trained on the same model, with the neurons of that model. We find that the SAE with the highest number of features to have the highest scores in the case of the Gemma 2 9b SAEs and that SAEs trained on the residual stream have higher scores for both Gemma 2 9b and Llama 3.1 8b.

The scorer and the explainer model were Llama 3.1 70b instruct, quantized to 4bit.

\begin{table}[t]
\caption{Comparison of SAEs for Fuzzing and Detection. Numbers shown are the median score and the interquartile (25\%-75\%) range.}
\label{tab:feature_size}
\centering
\begin{tabular}{lcc}
\toprule
Experiment & Fuzzing & Detection \\
\midrule
Random interpretation & 0.51 (0.45--0.57) & 0.51 (0.45--0.58) \\
Randomly initialized Topk SAE & 0.55 (0.50--0.60) & 0.54 (0.50--0.59)  \\
\addlinespace
131k features Gemma 2 9b & \textbf{0.76 (0.67--0.86)} & \textbf{0.74 (0.63--0.85)}  \\
16k features Gemma 2 9b & 0.73 (0.63--0.83) & 0.70 (0.59--0.79) \\
Top 32 neurons Gemma 2 9b & 0.62 (0.54--0.70) & 0.59 (0.53--0.64) \\
Top 256 neurons Gemma 2 9b & 0.59 (0.53--0.65) & 0.57 (0.52--0.62) \\
\addlinespace
262k features Llama 3.1 8b MLP & 0.79 (0.69--0.86) & 0.79 (0.64--0.85) \\
262k features Llama 3.1 8b & \textbf{0.81 (0.71--0.86)} & \textbf{0.83 (0.68--0.85)} \\
Top 32 neurons Llama 3.1 8b & 0.55 (0.51--0.62) & 0.53 (0.50--0.59) \\
Top 256 neurons Llama 3.1 8b & 0.54 (0.49--0.60) & 0.53 (0.50--0.57) \\
\bottomrule
\end{tabular}
\end{table}



\section{Automatically interpreting interventions}
\label{app:output}
\subsection{Scoring implementation}

For each feature, we sample a pool of 40 length-64 prompts from RPJv2~\citep{together2023redpajama}, of which 30 i.i.d. prompts are taken for scoring, while the remaining 10 are used by the explainer. The pool is sampled among nonzero activating contexts, stratified by the quintile of the feature's max activation in that context (i.e., the nonzero activating documents are sorted by the context's maximum activation of the current feature, then 8 contexts are sampled from the first quintile, 8 from the second etc.). Each of these prompts is then truncated to include only the first token on which the feature activates, so that all previous activations for the feature are 0. Activations on the first token position are ignored because the SAEs were not trained on that position.

We filter out features that activates on less than 200 of the 10 million RPJ-v2 tokens. 

We then sample generations with a maximum of 8 new tokens from the subject model. For generations with the intervention, we perform an additive intervention on the feature at all token positions after and including the final prompt token. We tune the intervention strength of each feature to various KL-divergence values on the scoring set with a binary search. We stop the binary search when the KL divergence is within 10\% of the desired value\footnote{Sometimes the KL-divergence is not perfectly monotonic in the intervention strength so 10\% error is exceeded. We report the average KL divergence that we observe in Figure~\ref{fig:intervention-scores-vs-layer}}. For our zero-ablation experiment, instead of doing an additive intervention we clamp the feature's activation to 0. Specifically, the hidden states are encoded by the SAE, then the SAE reconstruction error is computed using a clean decoding, then the SAE encoding is clamped and decoded, and the error is added to the clamped decoding.

The scorer is Llama-3.1 8B base. We use the base model for improved calibration, and prompt it as follows.
\begin{mdframed}
\begin{verbatim}
<PASSAGE>
from west to east, the westmost of the seven wonders of the 
world is the great wall of china

The above passage contains an amplified amount of "Asia"

<PASSAGE>
Given 4x is less than 10, 4

The above passage contains an amplified amount of "numbers"

<PASSAGE>
In information theory, the information content, self-
information, surprisal, or Shannon information is a basic 
quantity derived by her when she was a student at Windsor

The above passage contains an amplified amount of 
"she/her pronouns"

<PASSAGE>
My favorite food is oranges

The above passage contains an amplified amount of 
"fruits and vegetables"

<PASSAGE>
...
\end{verbatim}
\end{mdframed}

\subsection{Generating interpretations}

We generate interpretations using only the intervention's effect on the subject's next-token probabilities because this leads to a concise and precise prompt. The explainer, like the scorer, is Llama-3.1 8B.

The explainer sees a distribution of 10 prompts that is sampled i.i.d. from the same population as the scorer's prompts.

We use the following prompt for the explainer, with 3 few-shot examples truncated for brevity.
\begin{mdframed}
\begin{verbatim}
We're studying neurons in a transformer model. We want to know 
how intervening on them affects the model's output.

For each neuron, we'll show you a few prompts where we 
intervened on that neuron at the final token position, and the 
tokens whose logits increased the most.

The tokens are shown in descending order of their probability 
increase, given in parentheses. Your job is to give a short 
summary of what outputs the neuron promotes.

Neuron 1
<PROMPT>Given 4x is less than 10,</PROMPT>
Most increased tokens: ' 4' (+0.11), ' 10' (+0.04), 
' 40' (+0.02), ' 2' (+0.01)

<PROMPT>For some reason</PROMPT>
Most increased tokens: ' one' (+0.14), ' 1' (+0.01), 
' fr' (+0.01)

<PROMPT>insurance does not cover claims for accounts 
with</PROMPT>
Most increased tokens: ' one' (+0.1), ' more' (+0.02), 
' 10' (+0.01)

interpretation: numbers

Neuron 2
...
\end{verbatim}
\end{mdframed}

\subsection{Baselines}

\textbf{Random SAE}. We experiment with a random top-k SAE with $k = 50$, roughly the average sparsity of the gemma SAEs. The encoder is initialized with 131,072 spherically uniform unit-norm features, and the decoder is initialized to its transpose. We use a random SAE with TopK activations because we need sparsity for our sampling procedure to work properly.

\textbf{Random interpretations}.
For each layer and target KL value, we compute a random interpretation baseline where we shuffle the interpretations across features.

\subsection{Intervention interpretability results}
\begin{figure}
    \centering
    \includegraphics[width=\linewidth]{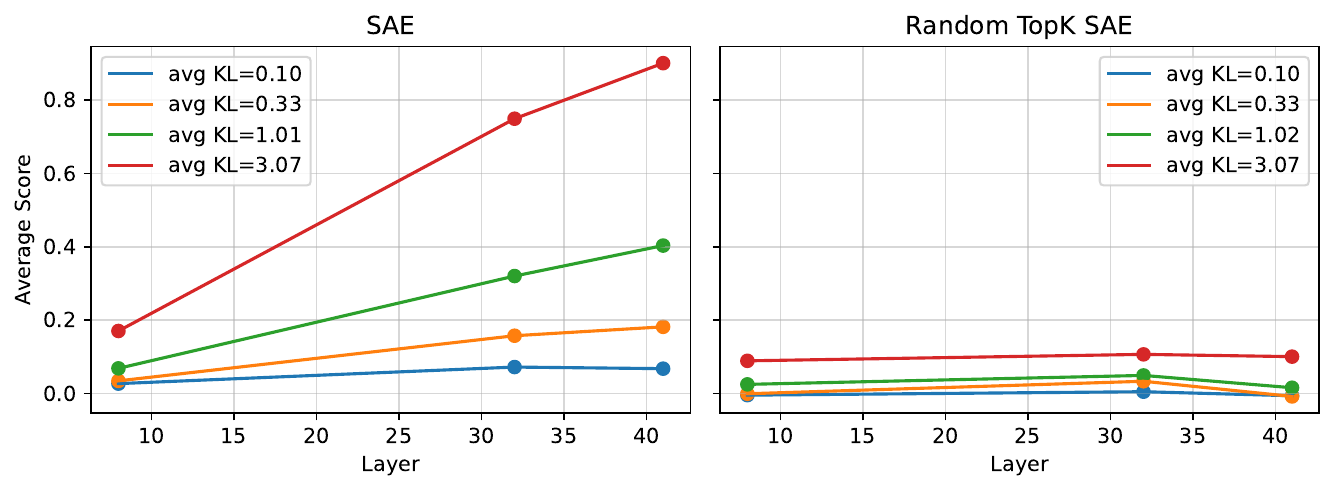}
    \caption{Average intervention score vs layer. SAE features in later layers have more explainable effects on output. Random features, however, have uninterpretable effects on output, even at late layers.}
    \label{fig:intervention-scores-vs-layer}
\end{figure}

Figure~\ref{fig:intervention-scores-vs-layer} shows the average intervention score at each layer, for a few KL values. Intervention scores are higher in later layers, likely because of the increased proximity to the model's output.

\begin{figure}
    \centering
    \includegraphics[width=0.99\linewidth]{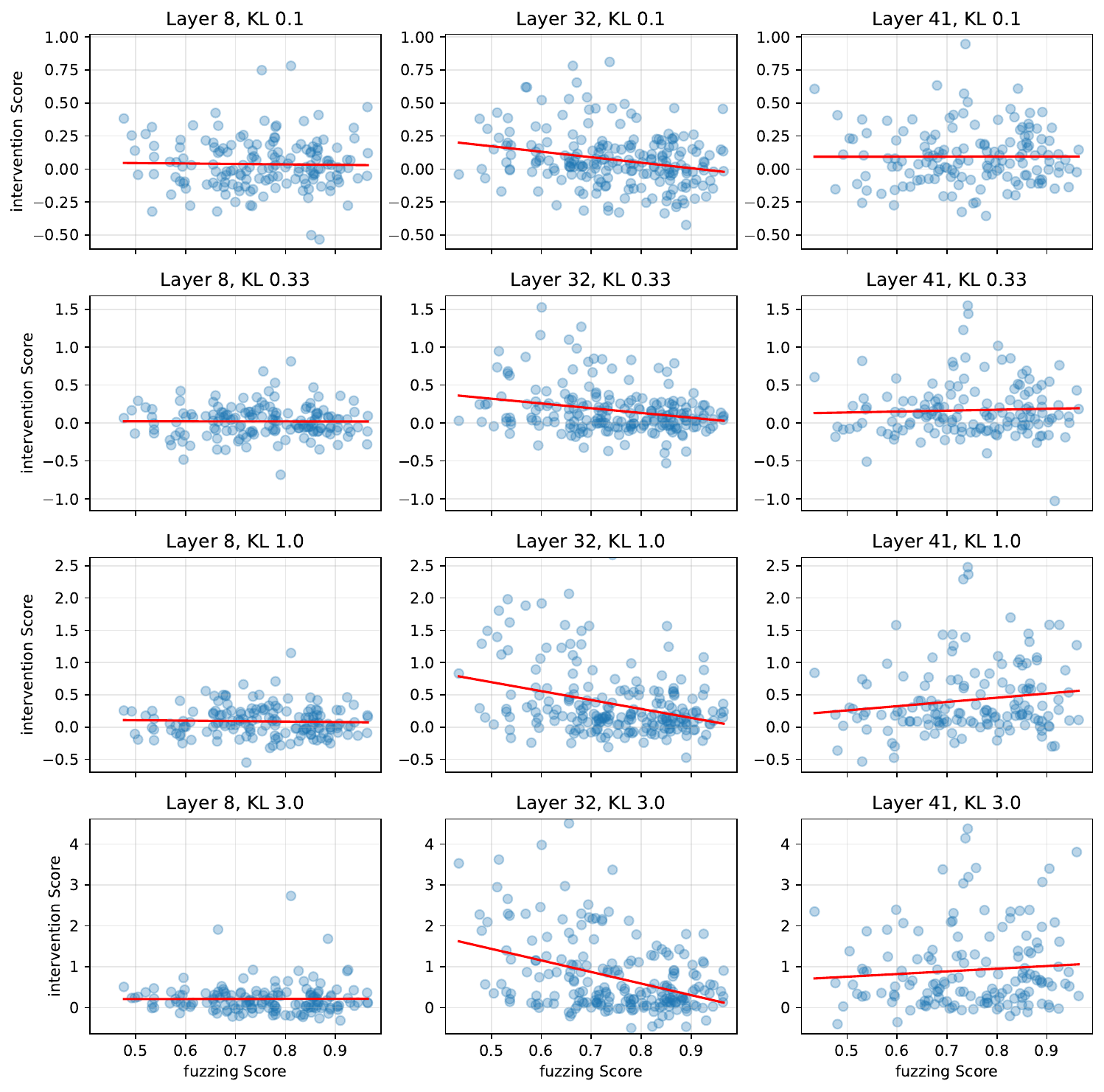}
    \caption{Comparison of intervention and fuzzing scores across layers at various target KL values.}
    \label{fig:intervention-vs-fuzzing}
\end{figure}

Figure~\ref{fig:intervention-vs-fuzzing} is similar to the right half of Figure~\ref{fig:intervention-hists-and-scatter}, but at multiple layers and KL values. Early layers have small intervention scores across the board, so there is little correlation, while the reason for little correlation in layer 41 is less clear. 

\end{document}